\documentclass[runningheads]{llncs}

\usepackage[mobile]{eccv}

\usepackage{eccvabbrv}

\usepackage{graphicx}
\usepackage{booktabs}

\usepackage{tikz}
\usepackage{comment}
\usepackage{amsmath,amssymb} % define this before the line numbering.
\usepackage{color}
\usepackage{array}
\usepackage{pifont}
\usepackage{multirow}

\usepackage{wrapfig}
\usepackage{lmodern}

\usepackage[accsupp]{axessibility}  % Improves PDF readability for those with disabilities.

\definecolor{whitegray}{rgb}{.45,.45,.45}

\newcommand{\gray}[1]{{\textcolor{whitegray}{#1}}}

\newcommand{\xmark}{\ding{55}}
\newcolumntype{Y}{>{\centering\arraybackslash}X}

\def\Plus{\texttt{+}}

\newif\ifCommentsAuthors
\CommentsAuthorsfalse
\ifCommentsAuthors

    \definecolor{myred}{rgb}{.8,.0,.0}

    \definecolor{myblue}{rgb}{0,0.4,.8}

    \definecolor{mcolor}{rgb}{0,0.5,0.1}

    \definecolor{mygreen}{rgb}{.0,.8,.0}

    \definecolor{hermancolor}{HTML}{FF6600}

\else
    \definecolor{myred}{rgb}{.8,.0,.0}

\fi

\usepackage{hyperref}

\usepackage{orcidlink}

\begin{document}

\title{UPose3D: Uncertainty-Aware 3D Human Pose Estimation with Cross-View and Temporal Cues}

\titlerunning{UPose3D: Uncertainty-Aware 3D Human Pose Estimation}

\author{Vandad Davoodnia
% \thanks{This work was performed during Vandad Davoodnia's internship at Ubisoft Laforge.}% <-this % stops a space
% \thanks{V. Davoodnia and A. Etemad were with the Department of Electrical and Computer Engineering, Queen's University, Kingston, Ontario, Canada. e-mail: \{vandad.davoodnia,ali.etemad\}@queensu.ca}% <-this % stops a space
% \thanks{V. Davoodnia and S. Ghorbani were with Ubisoft LaForge, Toronto, Ontario, Canada. e-mail: saeed.ghorbani@ubisoft.com}% <-this % stops a space
% \thanks{M.~A.~Carbonneau and A. Messier were with Ubisoft LaForge, Montreal, Québec, Canada. e-mail: \{marc-andre.carbonneau2, alexandre.messier\}@ubisoft.com}
\inst{1,2}\orcidlink{0000-0002-2167-2119} \and
Saeed Ghorbani\inst{2}\orcidlink{0000-0002-3227-9013} \and
Marc-André Carbonneau\inst{2}\orcidlink{0000-0002-0677-415X} \and
Alexandre Messier\inst{2} \and
Ali Etemad\inst{1}\orcidlink{0000-0001-7128-0220}}

\authorrunning{V.~Davoodnia et al.}

\institute{Queen's University, Canada  \and 
Ubisoft LaForge, Canada \\ 
\email{\{vandad.davoodnia,ali.etemad\}@queensu.ca} \\
\email{\{saeed.ghorbani,marc-andre.carbonneau2,alexandre.messier\}@ubisoft.com}
% \email{\{saeed.ghorbani,marc-andre.carbonneau2,alexandre.messier\}@ubisoft.com}
}

%******************
\maketitle
\begin{abstract} 
We introduce UPose3D, a novel approach for multi-view 3D human pose estimation, addressing challenges in accuracy and scalability. Our method advances existing pose estimation frameworks by improving robustness and flexibility without requiring direct 3D annotations. At the core of our method, a pose compiler module refines predictions from a 2D keypoints estimator that operates on a single image by leveraging temporal and cross-view information. Our novel cross-view fusion strategy is scalable to any number of cameras, while our synthetic data generation strategy ensures generalization across diverse actors, scenes, and viewpoints. Finally, UPose3D leverages the prediction uncertainty of both the 2D keypoint estimator and the pose compiler module. This provides robustness to outliers and noisy data, resulting in state-of-the-art performance in out-of-distribution settings. In addition, for in-distribution settings, UPose3D yields performance rivalling methods that rely on 3D annotated data while being the state-of-the-art among methods relying only on 2D supervision.
\end{abstract}

\section{Introduction}
Multi-view 3D human pose estimation is a challenging task in computer vision that involves determining the 3D position of human body landmarks given videos or images from multiple synchronized cameras \cite{usman2022metapose,iskakov2019learnable,he2020epipolar}. Compared to monocular setups, multi-view pose estimation leverages information from different viewpoints, alleviating the single-camera ambiguity and improving accuracy in challenging situations. This robustness is crucial in precision-demanding applications like markerless motion capture, essential to industries such as video gaming and film-making, where sub-centimeter accuracy is often required.

The conventional 3D keypoint estimation process involves two stages. Firstly, 2D landmarks are extracted from each camera view. This is followed by triangulation using the known camera parameters to infer 3D points \cite{iskakov2019learnable,zhang2021adafuse}. However, the accuracy of such methods heavily relies on the precision of independent 2D predictions across views, which is problematic in scenarios with complex body-part interactions or severe occlusions. Outlier mitigation techniques such as RANdom-SAmple Consensus (RANSAC) \cite{martinez2022ransac} and refinement algorithms \cite{usman2022metapose} offer some robustness but cannot fully address these inherent limitations. Recent advances in deep learning models that use cross-view fusion strategies \cite{shuai2022adaptive,qiu2019cross,he2020epipolar} have yielded promising 3D pose estimation results. For instance, Epipolar Transformers \cite{he2020epipolar} shows the benefits of leveraging cross-view information using epipolar geometry. However, the scalability of such approaches is often hindered with additional cameras, requiring advanced techniques to maximize the agreement between several model outputs \cite{qiu2019cross}. Another research direction aims to leverage rich temporal information \cite{tang20233d,shuai2022adaptive} to enhance pose estimation accuracy. For instance, numerous works show the impact of using large temporal context for monocular 3D pose estimation \cite{zhu2023motionbert,reddy2021tessetrack}. Similarly, methods like MFT-Transformers \cite{shuai2022adaptive}, focusing on multi-view fusion and temporal modelings, can yield improvements over single-frame methods. However, such approaches require access to large annotated 3D datasets (multi-view video streams and corresponding 3D pose coordinates) during training, which is especially scarce in outdoor and in-the-wild settings. Furthermore, these models are often trained on limited pose and camera variations, hindering their generalization to novel views.

To address the challenges of viewpoint scalability and reliance on 3D annotated training data, we introduce UPose3D, a new method for 3D human pose estimation. Our method leverages 2D keypoints and their uncertainties from two sources to improve robustness to outliers and noisy data. These sources include: \textit{a}) direct 2D pose estimation from RGB images, and \textit{b}) a pose compiler module that utilizes consistency across views and over time. Additionally, we introduce a new cross-view fusion strategy to ensure scalability to different numbers of cameras before our pose compiler. Specifically, we project the keypoints from all available views onto a reference view to obtain a 2D point cloud for each joint. These are then fed to a point cloud transformer module to learn cross-view representations. These features are then passed to a spatiotemporal encoder to efficiently process temporal and skeleton information from temporal windows. To train our pose compiler without relying on 3D annotated datasets, we generate synthetic data simulating realistic multi-view human pose recordings from a large-scale motion capture dataset. This approach promotes generalization across diverse camera configurations and postures, overcoming the limitations of real-world, multi-view 3D annotated datasets.

To evaluate the performance of our proposed model, we use four widely used public datasets, Human3.6m \cite{ionescu2013human3}, RICH \cite{huang2022capturing}, HUMBI \cite{yoon2021humbi,yu2020humbi}, and CMU Panoptic \cite{joo2015panoptic}, across two separate experiment setups. 
More specifically, we assess the performance of our method in both in-distribution (InD) and out-of-distribution (OoD) settings to better evaluate generalizability to new environments and multi-view camera setups. These experiments demonstrate that our approach outperforms prior state-of-the-art solutions in OoD while achieving competitive performance in InD settings. Next, we provide detailed ablation experiments demonstrating the impact of uncertainty modeling and our pose compiler in improving multi-view pose estimation robustness to outliers. Finally, we perform several experiments to showcase the impact of the number of camera views and larger time windows. In summary, our method achieves a high level of view-point flexibility and robustness without requiring direct 3D annotations. Our contributions can be summarized as follows:
\begin{itemize}
    \item We present UPose3D, a 3D human pose estimation pipeline for multi-view setups that achieves state-of-the-art results in OoD settings and performs competitively in InD evaluations.
    \item Our method uses a novel uncertainty-aware 3D pose estimation algorithm that uses normalizing flows to leverage 2D pose distribution modeling. Experiments demonstrate that this approach is more effective than prior works using heatmaps in terms of accuracy and computational costs. This also allows our pipeline to scale to different camera setups with \textit{constant} runtime.
    \item We propose a new training strategy that relies only on synthetic multi-view motion sequences generated online from motion capture data.
\end{itemize}

\section{Related Works}
\noindent\textbf{3D Pose Estimation.}
Traditionally, triangulation techniques like RANSAC have been used for 3D human pose estimation \cite{iskakov2019learnable,zhang2021adafuse}. However, these methods are generally not directly differentiable and their integration into deep learning pipelines is hindered. Therefore, recent research has explored more flexible, soft-predictive models, such as volumetric 3D keypoint representations \cite{iskakov2019learnable} and cross-view feature fusion strategies \cite{qiu2019cross}. Another approach \cite{bartol2022generalizable} introduces a stochastic framework for human pose triangulation that relies on 3D pose hypothesis generation, scoring, and selection from 2D detection of several camera views. However, the accuracy of 2D pose detectors limits most such approaches. As a result, some methods incorporate epipolar geometry for pose consistency via self-supervised \cite{kocabas2019self} and semi-supervised \cite{yao2019monet} learning. More advanced techniques have explored feature fusion strategies using epipolar lines across camera pairs \cite{he2020epipolar,zhang2021adafuse}, showing significant improvements. However, these methods require complex re-projection in view pairs, limiting their scalability to a large number of cameras or with different placements due to lack of variety in the training set. 

To avoid relying on large parameters, a recent work \cite{shuai2022adaptive} proposed a multi-view temporal transformer network for end-to-end 3D pose estimation by leveraging cross-view feature fusion. Other works have conducted experiments on using human pose priors using generative models such as GANs \cite{kocabas2020vibe} and Diffusion probabilistic models \cite{holmquist2023diffpose} to perform 3D pose estimations from a variety of inputs. However, most such approaches rely on 3D pose annotations from in-studio collected datasets to train their model. Another research track explores the benefits of uncertainty modeling for 3D human pose estimation for enhancing the performance in occlusion-heavy scenarios \cite{bramlage2023plausible,kundu2022uncertainty}. For example, Residual Log-likelihood Estimation (RLE) \cite{li2021human}, has been proposed to model the underlying distribution of 2D keypoints in human pose estimation via regression. This approach leverages a re-parameterization technique and normalizing flows to learn keypoint uncertainty and achieves similar performance to heatmap-based \cite{sun2019deep,cao2017realtime} and SimCC-based \cite{li2022simcc} techniques. We exploit the ability of RLE to estimate a differentiable likelihood of keypoints, which other methods lack.

\noindent\textbf{Transformers in Pose Estimation.} 
The Transformer architecture and its self-attention mechanism have significantly advanced Natural Language Processing and Computer Vision. Self-attention's ability to capture long-range dependencies makes it invaluable for 3D pose estimation, which requires careful consideration of spatial, temporal, and multi-view information. Recent works effectively leverage Transformers for 3D pose estimation in both single-camera setups \cite{lin2021end,zhu2021posegtac} and for handling spatiotemporal information within single images \cite{llopart2020liftformer,li2022mhformer,zheng20213d}. Additionally, transformers show promise in aggregating multi-view clues via epipolar geometry \cite{he2020epipolar,ma2021transfusion}. Despite the impressive performance of transformers in a variety of tasks, their memory requirements cause an obstacle in processing all spatial, temporal, and multi-view information together. As a result, a group of researchers has adopted a criss-cross attention mechanism to limit each attention layer's receptive field without sacrificing the network's overall receptive field \cite{huang2019ccnet}. Recent work has also adopted the criss-cross attentions \cite{tang20233d} to process temporal and joint information for human 3D pose estimation, showing superior performance compared to parallel or concurrent models.

\begin{figure}[t]
\centering
\includegraphics[width=0.95\textwidth]{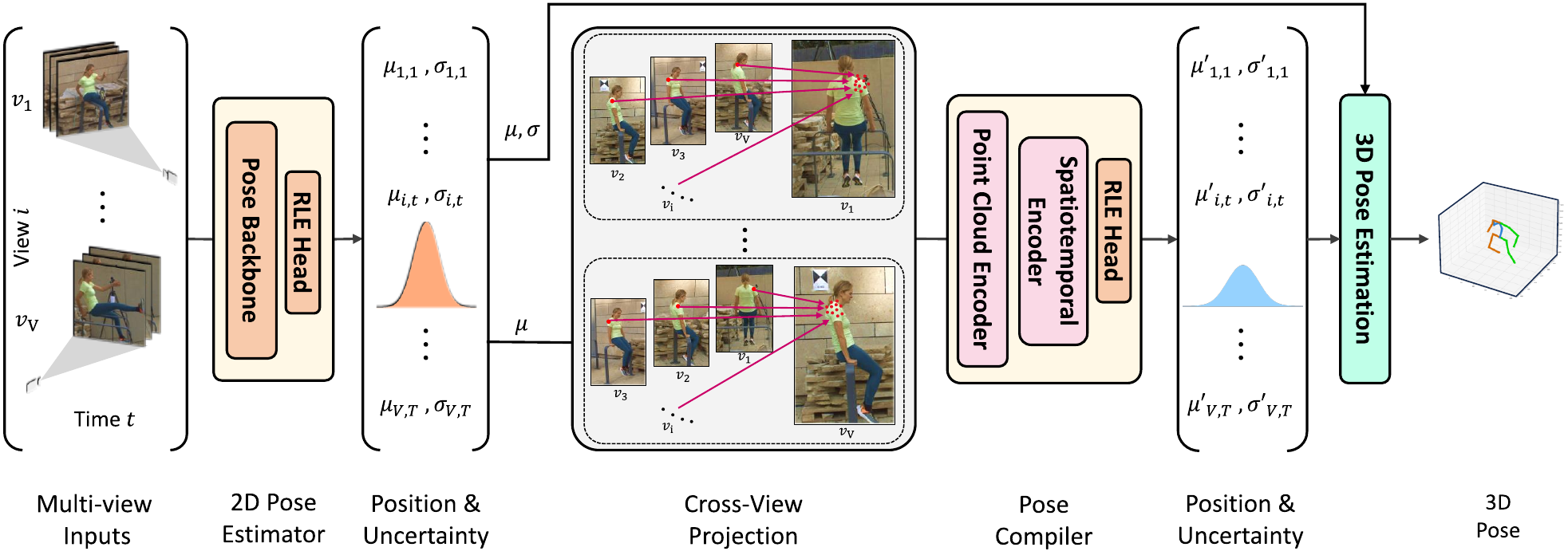}
\caption{We illustrate the key stages of UPose3D. It begins with extracting 2D keypoints and uncertainties from the multi-view videos. The keypoints are then projected onto each reference view using epipolar geometry. Our pose compiler is then used to refine the predictions by leveraging cross-view and spatiotemporal information. Finally, the 3D pose is obtained using the keypoint and uncertainty predictions of each stage.}
\label{fig:fig1}
\end{figure}

\section{Method} 
Upose3D determines the 3D coordinates of body joints from one or more consecutive multi-view frames. \Cref{fig:fig1} depicts an overview of our proposed method. At first, an RLE-based 2D pose estimator (\cref{sec:meth_1_2d}) extracts 2D keypoints and their corresponding uncertainties from the input images. The keypoints are then projected onto other views using epipolar geometry (\cref{sec:meth_2_mv}). From there, the pose compiler (\cref{sec:meth_3_mvf}) refines the keypoint locations and their uncertainties by leveraging spatiotemporal and cross-view information. Lastly, using outputs from the 2D pose estimator and the pose compiler, an iterative algorithm obtains the final 3D poses (\cref{sec:meth_5_opt}). We end this section by describing our multi-view training data synthesis (\cref{sec:meth_6_synth}).

\subsection{2D Pose Estimation} \label{sec:meth_1_2d}
The first step of our method is estimating 2D joint position in each image $\mathcal{I}_{i,t}$ from all camera views $i \in \mathcal{V} = \{1,2,\dots, V\}$ and frames $t \in \mathcal{T} = \{1,2,\dots, T\}$. Similar to \cite{li2023hybrik,dwivedi2023poco}, our method implements a single layer Residual Log-likelihood Estimation (RLE) head \cite{li2021human} on top of an off-the-shelf backbone (\eg, CPN \cite{chen2018cascaded}). Aside from being computationally effective and robust to occlusion, RLE provides uncertainty $\hat{\sigma}$ for each joint position prediction $\hat{\mu}$. Specifically, the RLE predicts a distribution $P_\Theta(x | \mathcal{I})$ that models the probability of the ground truth appearing in position $x$ using a normalizing flow model \cite{rezende2015variational} with learnable parameters $\Theta$. The $\hat{\mu}$ produced by this module is refined in the next stages by leveraging cross-view and temporal information. The estimated $\hat{\mu}$, $\hat{\sigma}$, along with the normalizing flow parameters $\Theta$, are used for 3D keypoint estimation in \cref{sec:meth_5_opt}.

\subsection{Cross-view Projection} \label{sec:meth_2_mv}
Next, we leverage the information from multiple camera views by projecting the 2D keypoints from one view to another using epipolar geometry. We derive a fundamental matrix $F_{ij} \in \mathbb{R}^{3 \times 3}$ relating two camera views $i,j \in \mathcal{V}$ from known intrinsic and extrinsic camera parameters. For each predicted keypoint $\hat{\mu}_j$ in view $j$, we obtain the epipolar line in the reference view $i$ by $I_{ij}=F_{ij}^\intercal \hat{\mu}_j$ \cite{hartley2003multiple}. Next, we find the closest point on $I_{ij}$ to the keypoint $\hat{\mu}_i$ in view $i$. This point represents the projection of the keypoint from view $j$ onto the reference view $i$. By repeating this cross-view projection for all available views $i$, joints $k \in \mathcal{J} = \{1,2,\dots, J\}$, and time frames $t$, we create 2D point clouds $\mathcal{C}_{i,k,t}$ containing the projected keypoint. Finally, for consistency between various scales and views, we normalize the elements of $\mathcal{C}$ using the subject's bounding box within their reference view.

\begin{figure}[t]
\centering
\includegraphics[width=0.95\textwidth]{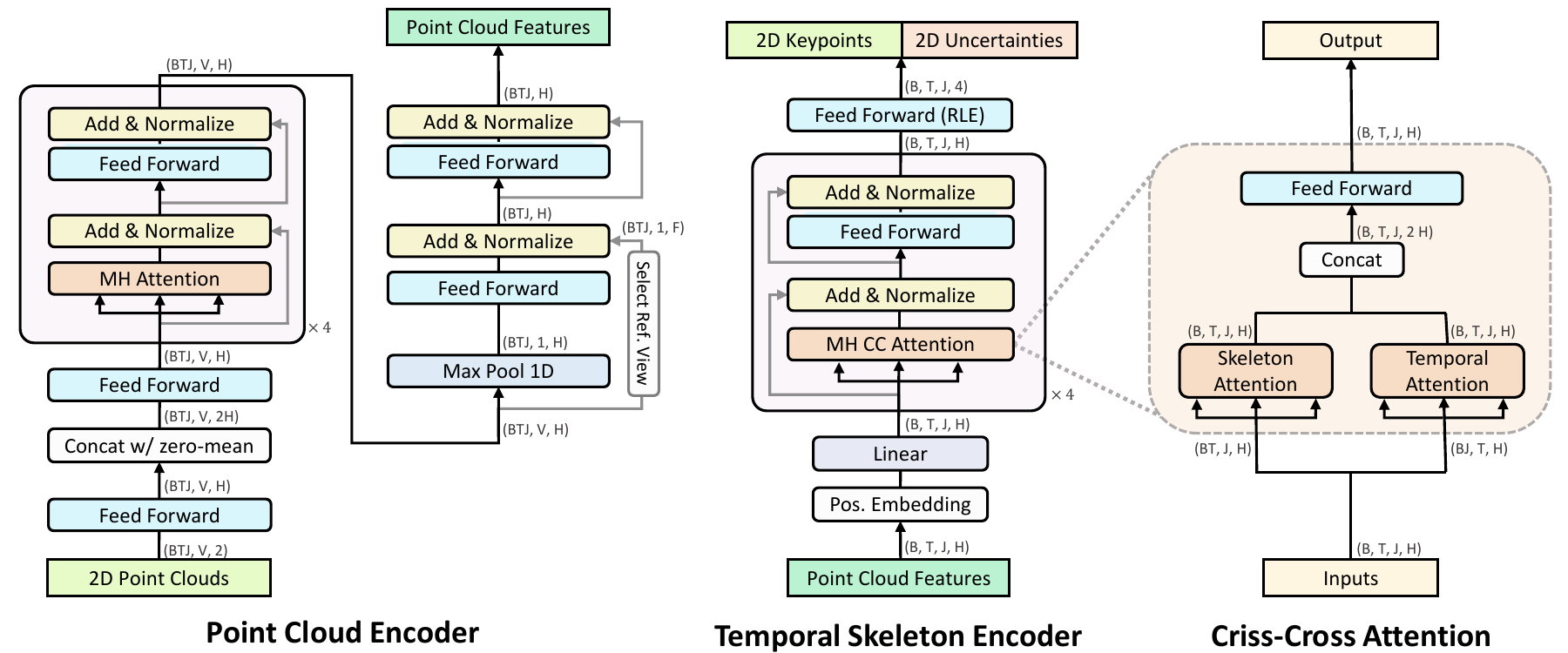}
\caption{Architecture of the proposed pose compiler module consisting of a point cloud encoder and a spatiotemporal encoder with criss-cross attention. Tensor sizes depend on the batch size $B$, temporal window length $T$, number of joints $J$, camera views $V$ and the point cloud feature dimensionality $H$.}
\label{fig:fig2}
\end{figure}

\subsection{Pose Compiler}  \label{sec:meth_3_mvf}
The pose compiler module aggregates multi-view information embedded in the 2D point clouds and leverages temporal information (\ie, joint positions across time). As shown in \cref{fig:fig1}, the pose compiler consists of 2 main parts: a point cloud encoder and a spatiotemporal encoder. The point cloud encoder first extracts a feature vector $f_{i,k,t}$ that describes a point cloud $\mathcal{C}_{i,k,t}$. Our implementation is inspired by the na\"ive Point Cloud Transformers \cite{guo2021pct,ren2022benchmarking}, but we modify its architecture to preserve coordinate information. Specifically, we use multi-head attention and a residual connection from the features of the reference view to the output of the last max-pooling layer. For the detailed architecture of our point cloud encoder, please refer to \cref{fig:fig2}.

Next, we concatenate the feature vectors from every frame and joint in each reference view $\{\textbf{f}_{i}\}$ with temporal and spatial position embeddings, \ie, frame and skeleton joint ID. The results are then passed to our spatiotemporal encoder that implements an RLE head to provide position $\hat{\mu}'_{i,k,t}$ and uncertainty $\hat{\sigma}'_{i,k,t}$ estimations. Our spatiotemporal encoder is a transformer \cite{vaswani2017attention}, with specific attention block modifications that accommodate the large dimensionality of the tensors containing all temporal and skeleton information. As detailed in \cref{fig:fig2}, we replace the standard attention module with criss-cross attentions \cite{huang2019ccnet,tang20233d}, which approximates full temporal and spatial dependencies while being more memory efficient than full attention. We train the RLE head of the spatiotemporal encoder using the same strategy as \cite{li2021human}. Finally, similarly to \cref{sec:meth_1_2d}, the RLE head simultaneously maximizes and learns a distribution $P_{\Theta'}(x | \mathcal{C})$ that represents the probability of the appearance of a keypoint in position $x$ using a normalizing flow model with learnable parameters $\Theta'$. In the next section, we use the estimated $\hat{\mu}'$, $\hat{\sigma}'$, and $\Theta'$ for 3D pose estimation.

\subsection{3D Pose Estimation} \label{sec:meth_5_opt}
As the final step, our method obtains the 3D keypoints using position estimates from 2D frames ($\hat{\mu}$, $\hat{\sigma}$), as well as refined versions from the pose compiler ($\hat{\mu}'$, $\hat{\sigma}'$). 
Specifically, we use the estimated keypoints as labels when sampling from the associated keypoint density functions ($P_{\Theta}(x | \mathcal{I})$ and $P_{\Theta'}(x | \mathcal{C})$) during Maximum Likelihood Estimation (MLE). Therefore, the total loss function of MLE for each joint and each time frame is defined as:
\begin{equation} \label{equ:equ1} \scriptstyle
    \mathcal{L}_{mle} = - \log \prod_{i \in \mathcal{V}}^{} P_\Theta(u_i | \mathcal{I}) \bigg |_{u_i=\hat{\mu}_i} - \log \prod_{i \in \mathcal{V}}^{} P_{\Theta'}(u_i | \mathcal{C}) \bigg |_{u_i=\hat{\mu}_i'},
\end{equation}
where $u_i$ is the projection of variable 3D point $U$ onto each camera view $i$. Minimizing this loss function increases the likelihood of $U$ appearing close to the ground-truth 3D keypoints $U_g$ without 3D supervision. To solve this non-convex optimization problem, we initialize $U$ with a Direct Linear Transformation (DLT) algorithm \cite{hartley2003multiple} and iteratively refine it using an optimizer, \ie, L-BFGS \cite{liu1989limited}.

\subsection{Training and Multi-view Data Synthesis} \label{sec:meth_6_synth}
Our proposed pose compiler does not require 3D annotated data, given that it operates on point clouds. Therefore, we can use any animation data to create a synthetic training set. The synthesis of multi-view training data starts by randomly positioning several cameras in 3D space to look roughly at the human body's center. During this process, we limit the camera height, distance, tilt, pitch, and yaw within a representative range of a normal multi-view recording setup. Next, we extract the 3D landmarks and project them onto each camera to obtain ground-truth 2D keypoints $\mu_g$. We then corrupt these keypoints through extensive augmentations. Refer to our Supplementary Materials for more details on our data synthesis. Finally, we obtain the point clouds training data from the corrupted 2D keypoint using the cross-view projection from \cref{sec:meth_2_mv}.

\section{Experiments}
\subsection{Datasets}
We compare our proposed approach with prior works on the Human3.6m \cite{ionescu2013human3} dataset and measure how our method generalizes to different skeleton configurations and unseen outdoor environments on the RICH \cite{huang2022capturing} dataset. We use the COCO WholeBody \cite{jin2020whole,lin2014microsoft} and MPII \cite{andriluka20142d} for training our 2D pose estimator in different scenarios (refer to the Supplementary Materials for more information \cite{zhou2017towards}). The pose compiler is trained using motion capture data from the AMASS dataset \cite{mahmood2019amass} to simulate multi-view training data. Finally, we analyze the viewpoint scalability of our approach on the CMU Panoptic \cite{joo2015panoptic} and HUMBI \cite{yoon2021humbi,yu2020humbi} datasets. The details of these datasets are as follows.

\noindent\textbf{Human3.6m.}
The Human3.6m \cite{ionescu2013human3} dataset is a standard benchmark for evaluating the performance of 3D human pose estimation solutions in multi-view \cite{shuai2022adaptive} and single-view \cite{zhu2023motionbert} settings. We report our InD performance in Protocol-I using subjects 1, 5, 6, 7, and 8 for training and 9 and 11 for testing.

\noindent\textbf{RICH.}
Real scenes, Interaction, Contact, and Humans (RICH) dataset \cite{huang2022capturing} is a dataset of multi-view videos with accurate bodies and 3D-scanned scenes obtained using markerless motion capture technology. We extract 3D ground-truth keypoints from the provided SMPL-x \cite{pavlakos2019expressive} bodies for evaluation. We report OoD performance on the test set, which contains 7 subjects in 52 scenarios and 3 environments, including a construction site, a gym, and a lecture hall.

\noindent\textbf{CMU Panoptic.}
The CMU Panoptic \cite{joo2015panoptic} dataset is a large collection of human body poses and interactions recorded from multiple views. The dataset provides videos from 480 VGA and 30 HD cameras and ground-truth 3D poses obtained from markerless 2D pose estimation and triangulation. We use a small subset of the validation set in line with \cite{iskakov2019learnable} to analyze UPose3D's camera scalability.

\noindent\textbf{HUMBI.}
The HUman Multiview Behavioral Imaging (HUMBI) \cite{yoon2021humbi,yu2020humbi} dataset is a large-scale multi-view dataset designed for modeling and reverse-rendering of the human body in different expressions. It contains 107 HD recordings of 772 subjects with natural clothing. We use this dataset only for OoD evaluations during our scalability analysis. Specifically, we evaluate on the first 80 subjects by extracting ground-truth 3D keypoints from the provided SMPL \cite{loper2015smpl} parameters.

% An important contribution of this dataset is the unification of 3D human bodies under the SMPL \cite{loper2015smpl} and SMPL-x \cite{pavlakos2019expressive} parametric models used to train pose and motion priors such as VPoser \cite{pavlakos2019expressive} and HuMoR \cite{rempe2021humor}. 
\noindent\textbf{AMASS.} 
The Archive of Motion Capture as Surface Shapes (AMASS) \cite{mahmood2019amass} is a large collection of 3D human motion capture datasets. It contains over 40 hours of recording from 300 subjects spanning 11,000 actions. We only use the training set of this dataset following prior works \cite{rempe2021humor,pavlakos2019expressive,davoodnia2024skelformer}. Since the extraction of landmarks requires a specialized joint regressor for each skeleton configuration, in our InD experiments, we use the SMPL\Plus H \cite{loper2015smpl} body model and a 17-joint regressor provided by \cite{moon2022neuralannot}. For our OoD experiments, we use the SMPL-x \cite{pavlakos2019expressive} body model and a whole-body 133-joint regressor from  \cite{davoodnia2024skelformer}.

\subsection{Evaluation Metrics}
We adopt the standard evaluation metrics from prior works for our InD experiments on the Human3.6m \cite{ionescu2013human3} dataset. Mean Per Joint Position Error (MPJPE) represents the distance between the ground-truth and predicted pose in millimeters, while its Procrustes Aligned version (PA-MPJPE) and the Normalized variant (N-MPJPE) show how well the predicted pose fits the ground-truth keypoints after a similarity transformation. Next, we report the Translation Aligned error (TA-MPJPE) and PA-MPJPE for our OoD experiments on the RICH \cite{huang2022capturing} dataset following prior works \cite{li2022cliff,kocabas2021pare,kolotouros2019learning,lin2021end,tripathi20233d,shen2023learning}.

\subsection{Implementation Details}
We implement our pipeline using PyTorch \cite{paszke2019pytorch} and MMPose \cite{mmpose2020}. Here, we provide the details of our neural networks and the computation costs of our pose compiler. We encourage readers to refer to our Supplementary Materials for information on our multi-view data synthesis and augmentation strategies.

To compare our method with prior works, we choose a CPN \cite{chen2018cascaded} backbone based on ResNet152 \cite{he2016deep} with an input size  384 $\times$ 384 and an RLE head \cite{li2021human} for 2D pose estimation. We fine-tune our 2D pose estimator jointly on Human3.6m \cite{ionescu2013human3} and MPII \cite{andriluka20142d} datasets starting from the weights of the checkpoints provided by \cite{iskakov2019learnable}. For our OoD experiments, we use an HRNet-W48 \cite{sun2019deep} with 384 $\times$ 288 input size and train it from scratch. 
Our point cloud encoder uses 4 multi-head self-attention blocks, each with 4 attention heads. Additionally, we reduce its hidden dimension to 64 to accommodate the relatively smaller size of our point clouds. Following the point cloud encoder, the features are concatenated with two positional embeddings of size 64 to represent the time frames and different skeleton joints. Our spatiotemporal encoder employs four criss-cross transformer blocks \cite{huang2019ccnet,tang20233d} with 4 attention heads each. We set the hidden layer dimension of our transformers to 64 and the hidden dimension of prediction heads to 1024. We then train our pose compiler module using AdamW optimizer \cite{loshchilov2017decoupled} with a batch size of 64 on the AMASS \cite{mahmood2019amass} dataset for both OoD and InD evaluations. We use a learning rate of 4e-5 with a warm-up factor of 1e-4 for the first 500 iterations and a cosine annealing scheduler over the next 20,000 iterations. The network training takes about 6 hours on an NVIDIA 2080 RTX GPU.

Unlike prior works that rely on rendering techniques for synthetic data generation \cite{gong2022progressive,zhang2021adafuse}, we generate samples online with varying numbers of camera views (up to 8) during training without significant computation overhead. Furthermore, we increase the diversity of the motion capture data by applying several augmentations to the motion capture data before 3D keypoint extraction, such as random 180-degree rotations and mirroring around the mid-hip point.

\subsection{Baselines}
On Human3.6m \cite{ionescu2013human3}, we compare our UPose3D with methods that infer 3D human keypoints using only 2D supervised models \cite{chen2018cascaded,he2020epipolar,qiu2019cross,usman2022metapose,karashchuk2021anipose,kocabas2019self}. Additionally, we provide a summary of methods that rely on direct 3D annotations \cite{jiang2023probabilistic,gordon2022flex,shuai2022adaptive,zhang2021adafuse,remelli2020lightweight,iskakov2019learnable}, except for a few \cite{jiang2023probabilistic,gordon2022flex,reddy2021tessetrack} that use uncalibrated cameras. Moreover, we compare our work with methods that only rely on multi-view 2D keypoints along with the camera parameters \cite{zhao2024triangulation,ma2021transfusion,he2020epipolar,qiu2019cross}. Please refer to our Supplementary Materials for a more in-depth comparison with weakly-supervised and semi-supervised approaches \cite{rhodin2018learning,kocabas2019self,wandt2021canonpose,gong2022progressive,sun2023bkind}, multi-view methods without that do not rely on camera parameters \cite{jiang2023probabilistic,gordon2022flex,reddy2021tessetrack}, and monocular 3D pose estimation methods \cite{shen2023learning,tripathi20233d,lin2021end,kolotouros2019learning,kocabas2021pare,li2022cliff}. We also report and compare the performance of our method against triangulation approaches, such as DLT, RANSAC, and RPSM \cite{qiu2019cross}. On the RICH dataset \cite{huang2022capturing}, we compare our model with triangulation approaches and replicate the performance of AdaFuse \cite{zhang2021adafuse}, which is one of the top-performing methods on the Human3.6m dataset \cite{ionescu2013human3}.

\begin{table}[t]
  \centering
  \caption{The comparison of our method in InD settings against prior multi-view works on the full test set of the Human3.6m dataset. The errors are reported in \textit{mm}.}
  \label{tab:tab1_h36m}
  \setlength
  \tabcolsep{3pt}
  \scriptsize
  \begin{tabular}{lccccc}
    \hline    
    \textbf{Method} & \textbf{Backbone} & \textbf{Frames} & \textbf{MPJPE}{\scriptsize$\downarrow$} & \textbf{PA-MPJPE}{\scriptsize$\downarrow$}  & \textbf{N-MPJPE}{\scriptsize$\downarrow$}  \\
    \hline\hline
    \multicolumn{6}{c}{\textit{\gray{3D Supervision}}} \\
    \hline
    \gray{Learnable Triangulation \cite{iskakov2019learnable}}    & \gray{ResNet152}  & \gray{1}   & \gray{20.7}    & \gray{17.0} & \gray{-}    \\
    \gray{Canonical Fusion \cite{remelli2020lightweight}}         & \gray{ResNet152}  & \gray{1}   & \gray{21.0}    & \gray{-}    & \gray{-}    \\ 
    \gray{AdaFuse \cite{zhang2021adafuse}}                        & \gray{ResNet152}  & \gray{1}   & \gray{19.5}    & \gray{-}    & \gray{-}    \\
    \gray{TesseTrack \cite{reddy2021tessetrack}}    & \gray{HRNet \cite{sun2019deep}} & \gray{5}   & \gray{18.7}    & \gray{-}    & \gray{-}    \\  
    \gray{MTF-Transformer\Plus \cite{shuai2022adaptive}}          & \gray{CPN}        & \gray{1}   & \gray{26.5}    & \gray{-}    & \gray{-}    \\
    \gray{MTF-Transformer\Plus \cite{shuai2022adaptive}}          & \gray{CPN}        & \gray{27}  & \gray{25.8}    & \gray{-}    & \gray{-}    \\ 
    \gray{Flex \cite{gordon2022flex}}                             & \gray{ResNet152}  & \gray{All} & \gray{30.2}    & \gray{-}    & \gray{-}    \\  
    \gray{Jiang \etal \cite{jiang2023probabilistic}}              & \gray{ResNet152}  & \gray{1}   & \gray{27.8}    & \gray{-}    & \gray{-}    \\  
    \hline\hline
    \multicolumn{6}{c}{\textit{2D Supervision}} \\
    \hline
    EpipolarPose \cite{kocabas2019self}         &  ResNet50   & 1   & 55.08 & 47.91 & 54.90     \\  
    AniPose \cite{usman2022metapose,karashchuk2021anipose}    & GT  & 1   & -     & 75.0    & 103.0 \\  
    MetaPose \cite{usman2022metapose}           & PoseNet \cite{kendall2015posenet} & 1   & - & 32.0  & 49.0   \\  
    DLT \cite{qiu2019cross}                     & ResNet152   & 1   & 36.3  & -     & -    \\
    DLT  \cite{chen2018cascaded}                & CPN         & 1   & 30.5  & 27.6  & 29.8 \\
    UPose3D (Ours)                              & ResNet152   & 1   & 31.0  & 29.0  & 31.2 \\
    UPose3D (Ours)                              & ResNet152   & 27  & 29.9  & 27.2  & 29.8 \\
    UPose3D (Ours)                              & CPN         & 1   & 26.9  & 24.1  & 26.2 \\
    UPose3D (Ours)                              & CPN         & 27  & 26.4  & 23.4  & 25.6 \\
    \hline
    \hline
    \multicolumn{6}{c}{\textit{2D Supervision \Plus~Camera Parameters}} \\
    \hline
    TransFusion \cite{ma2021transfusion}          & ResNet50    & 1   & 25.8    & -     & -    \\ 
    DLT\textsuperscript{*} \cite{he2016deep}      & ResNet50    & 1   & 71.2    & 62.8  & 70.9 \\
    UPose3D (Ours)\textsuperscript{*}             & ResNet50    & 27  & 57.3    & 51.7  & 57.2 \\
    Cross-view Fusion \cite{qiu2019cross}         & ResNet152   & 1   & 26.2    & -     & -    \\
    Epipolar Transformers \cite{he2020epipolar}   & ResNet152   & 1   & 19.0    & -     & -    \\ 
    TR Loss\textsuperscript{\ding{61}} \cite{zhao2024triangulation} & HRNet \cite{sun2019deep}    & 1     &  25.8  & - & - \\ 
    UPose3D (Ours)                                & CPN         & 27  & 26.2    & 23.1  & 25.4 \\
    \hline
    \multicolumn{6}{l}{- denotes the error was not reported in the original work.} \\
    \multicolumn{6}{l}{* denotes training from scratch with no additional data.} \\
    \multicolumn{6}{l}{\ding{61} denotes using 1\% of unlabeled testing data during training.} 
  \end{tabular}
\end{table}

\section{Results}
\subsection{In-Distribution Performance}
In \cref{tab:tab1_h36m}, we present a comparison of our InD results with other state-of-the-art methods on the Human3.6m dataset, divided into three categories: \textit{a}) methods that, unlike ours, are trained directly with 3D annotated data; \textit{b}) methods that leverage only 2D supervision; and \textit{c}) methods that rely on 2D supervision along with the camera parameters. We observe that UPose3D outperforms all other methods that rely on 2D supervision. In addition, it yields results that are competitive to state-of-the-art methods relying on 3D supervision. For example, our method achieves similar results to MTF-Transformers \cite{shuai2022adaptive}, which, like our pose compiler, takes in multi-view and temporal 2D keypoints as input but estimates the 3D poses directly using a deep network with 10.1\textit{M} parameters. Furthermore, among methods that use the camera parameters of the Human3.6m dataset, our method performs similarly to TR Loss \cite{zhao2024triangulation}, which uses 1\% of the unlabeled \textit{testing} set during training. Although Epipolar Transformers \cite{he2020epipolar} perform better than ours due to their effective feature fusion strategy, it relies on precise camera parameters and has constraints on the viewing angle of neighbouring cameras. The consistent performance of our approach in both groups indicates the effectiveness of our proposed pose compiler and training strategies in obtaining precise keypoints regardless of the target dataset's camera setup. 

We also show that the choice of the 2D keypoint estimator backbone is an important factor for accuracy and that CPN is preferred over ResNet152. Interestingly, a vanilla triangulation algorithm, DLT, can outperform prior works given accurate 2D predictions from strong pose estimators. Combining strong pose estimators with our method yields even better performance. For instance, UPose3D improves the MPJPE by 3.6 \textit{mm} compared to vanilla triangulation using a CPN backbone (DLT\Plus CPN). Additionally, we experiment with a ResNet-50 model trained from scratch in the third group (2D supervised with camera parameters) and observe a 19.5\% improvement when UPose3D is used. Finally, we demonstrate the positive impact of the temporal frame window, which reduces the MPJPE by 0.5 \textit{mm} when using 27 frames instead of a single frame.

\subsection{Out-of-Distribution Generalization}
To evaluate the performance of UPose3D in OoD settings, we compare it to the best-performing baseline on the Human3.6m \cite{ionescu2013human3} dataset, namely AdaFuse \cite{zhang2021adafuse}. Additionally, we report the results for triangulation techniques applied to 2D pose estimators in \cref{tab:tab2_rich}. In this experiment setup, neither the models nor their components are trained on the RICH \cite{huang2022capturing} dataset. First, we observe that our method, despite the training source dataset, obtains the best results by achieving half of the error of the next best method. Additionally, we observe that AdaFuse performs poorly on this dataset despite being considered state-of-the-art on the Human3.6m \cite{ionescu2013human3} dataset. Our analysis of AdaFuse performance on different clips of RICH \cite{huang2022capturing} indicates that this method performs well only if all the viewpoints' predictions are within a reasonable range, whereas occasional noisy predictions cause large triangulation errors. In contrast, our solution deduces the correct pose by effectively discarding views with low confidence. In conclusion, when training on in-studio datasets, methods such as AdaFuse fail to adapt to in-the-wild environments. At the same time, our approach uses uncertainty modeling to better generalize to unseen camera configurations. UPose3D is not limited to the source dataset skeleton configuration because synthetic data are used for training, and it performs consistently between InD and OoD evaluations.

\begin{table}[t]
\centering
  \caption{Comparison of our method in OoD setting on the RICH \cite{huang2022capturing} dataset against prior works. The training sources are denoted as (2D pose estimator, Pose compiler).}
  \label{tab:tab2_rich}
  \setlength
  \tabcolsep{4pt}
  \scriptsize
  \begin{tabular}{llcc}
  \hline
    \textbf{Method} & \textbf{Training sources} &  \textbf{MPJPE}$\downarrow$  &\textbf{PA-MPJPE}$\downarrow$  \\
    \hline\hline
    AdaFuse\textsuperscript{*} \cite{zhang2021adafuse}   & (Human3.6m\Plus MPII, Human3.6m) & 524.0  & 85.8   \\ 
    Ours ($T=27$)                      & (Human3.6m\Plus MPII, Human3.6m) & 51.8   & 43.6   \\ 
    Ours ($T=1$)                       & (Human3.6m\Plus MPII, AMASS)     & 49.9   & 42.3   \\ 
    
    \hline
    HRNet-W48\Plus DLT\textsuperscript{*}                & (COCO, N/A)                        & 66.0   & 55.1    \\ 
    HRNet-W48\Plus Grid Search\textsuperscript{*}        & (COCO, N/A)                        & 64.4   & 54.9    \\ 
    Ours ($T=1$)                       & (COCO, AMASS)                    & 36.2   & 33.4    \\ 
    Ours ($T=27$)                      & (COCO, AMASS)                    & 34.7   & 32.0    \\ 
    \hline 
    \multicolumn{4}{l}{* denotes our implementation of prior works. } \\
  \end{tabular}
\end{table}

\subsection{Ablation Study} \label{sec:sec_ablation}
To investigate the impact of each component, we systematically remove them during \textit{inference} and report the results with a CPN backbone on the test set of Human3.6m \cite{ionescu2013human3}. We perform all experiments using a model with a temporal window of 27 frames. Please refer to our Supplementary Materials for additional analysis on temporal length, spatiotemporal encoder's architecture, different point cloud formulations, and initialization strategies for 3D estimation.

\noindent\textbf{Pose Compiler.}
We employ the pose compiler module to improve the keypoint and uncertainty predictions using cross-view and spatiotemporal information. By ablating this component, we use the original keypoints and uncertainties predicted by our 2D pose estimator to perform MLE. As shown in \cref{tab:tab3_ablation}, this experiment results in an additional 9.28 \textit{mm} error resulting in a higher error compared to the DLT algorithm (see \cref{tab:tab1_h36m}). This shows the effectiveness of our pose compiler for 3D pose estimation using normalizing flows.

\begin{table}[t]
\centering
  \caption{Ablation experiments on the Human3.6m dataset with $T=27$.
  }
  \label{tab:tab3_ablation}
  \setlength
  \tabcolsep{4pt}
  \scriptsize
  \begin{tabular}{lcc}
    \hline    
    \textbf{Method} & \textbf{MPJPE}$\downarrow$  &\textbf{PA-MPJPE}$\downarrow$  \\
    \hline\hline
    \textbf{UPose3D}                     & 26.42 & 23.42   \\
    \ \ \ \ w/o compiler                 & 37.14 & 33.90   \\
    \ \ \ \ w/o image branch             & 69.90 & 50.97   \\
    \ \ \ \  w/o compiler uncertainty    & 26.42 & 23.58   \\
    \ \ \ \  w/o image uncertainty       & 27.61 & 24.88   \\
    \ \ \ \  w/o uncertainty             & 48.11 & 41.20   \\
    \ \ \ \ \ \ \ \ w/o image branch     & 77.25 & 54.02   \\
    \ \ \ \ \ \ \ \ w/o compiler         & 30.48 & 27.63   \\
    \hline
  \end{tabular}
\end{table}

\noindent\textbf{Image Branch.}
Removing the image branch from the pipeline causes a significant rise in the error, seen in \cref{tab:tab1_h36m}, indicating the significance of keeping the original predictions for the final estimation.

\noindent\textbf{Uncertainty Modeling.}
We employ the normalizing flows of our RLE heads during the MLE loss minimization stage to incorporate the uncertainties of our predictions toward improving the model's robustness. 
We employ the RLE head's normalizing flows module to model the uncertainties, improving robustness. To study the effect of uncertainty modeling, we remove the RLE heads from the 2D pose estimator and pose compiler branches. Accordingly, we first remove the uncertainties from our pose compiler and replace our likelihood loss function (\cref{equ:equ1}) with a reprojection distance loss. \Cref{tab:tab1_h36m} shows that the PA-MPJPE error slightly rises, but the MPJPE remains unchanged. However, it should be noted that removing the pose compiler uncertainty during \textit{training} causes more performance degradation by resulting in 26.8 \textit{mm}, which shows the significance of uncertainty modeling. Next, we remove the image pose estimator uncertainties, which results in a 1.2 \textit{mm} rise in the error. Finally, we completely remove the uncertainties from both branches, effectively reducing our problem to a classic triangulation problem without confidence that can be solved via the DLT algorithm. This variation shows over 20 \textit{mm} higher error than UPose3D. In our final two experiments, we analyze the image and pose compiler predictions without uncertainty modeling. The first experiment shows the highest error, while the second experiment results in the same performance as a simple DLT algorithm without cross-view fusion.

\subsection{Computation Costs}
In \cref{tab:tab4_computation}, we compare the total computation costs of our method in comparison to state-of-the-art 3D approaches in single-frame scenarios. Compared to others, UPose3D has fewer parameters and significantly less computational cost, especially with more cameras. The computational cost of our 10-frame model can be broken down to 508.5\textit{G} FLOPs for CPN, 0.385\textit{G} for pose compiler, and 8.8$\pm$1.25\textit{G} for the optimization process. Due to the computational cost fluctuations caused by the L-BFGS optimizer, we measure it by averaging 20 runs using randomly selected cameras from the HUMBI dataset \cite{yoon2021humbi,yu2020humbi}.

\begin{table}[t]
\centering
\centering
  \setlength\tabcolsep{3pt}  
  \scriptsize
  \caption{Comparison of computation costs.}
  \begin{tabular}{lccc}
    \hline
    Method & Params{\scriptsize(\textit{M})$\downarrow$} & FLOPs for 4 Cams{\scriptsize(\textit{G})$\downarrow$} & FLOPs for 10 Cams{\scriptsize(\textit{G})$\downarrow$} \\
    \hline
    \hline
    Learnable Triangulation \cite{iskakov2019learnable} & 80.6 & 716.1 & 1326.9  \\ 
    Epipolar Transformers \cite{he2020epipolar}         & 68.1 & 406.5 & 1016.2  \\ 
    MTF-Transformers \cite{shuai2022adaptive}           & 78.6 & 407.0 & 1017.8  \\  
    AdaFuse \cite{zhang2021adafuse}                     & 69.7 & 595.0 & 1487.6  \\ 
    Ours                                                & 65.4 & 208.7 &  517.7  \\ 
    \hline 
  \end{tabular}
  \label{tab:tab4_computation}
\end{table}

\begin{figure}[t]
\centering
\includegraphics[width=0.95\textwidth]{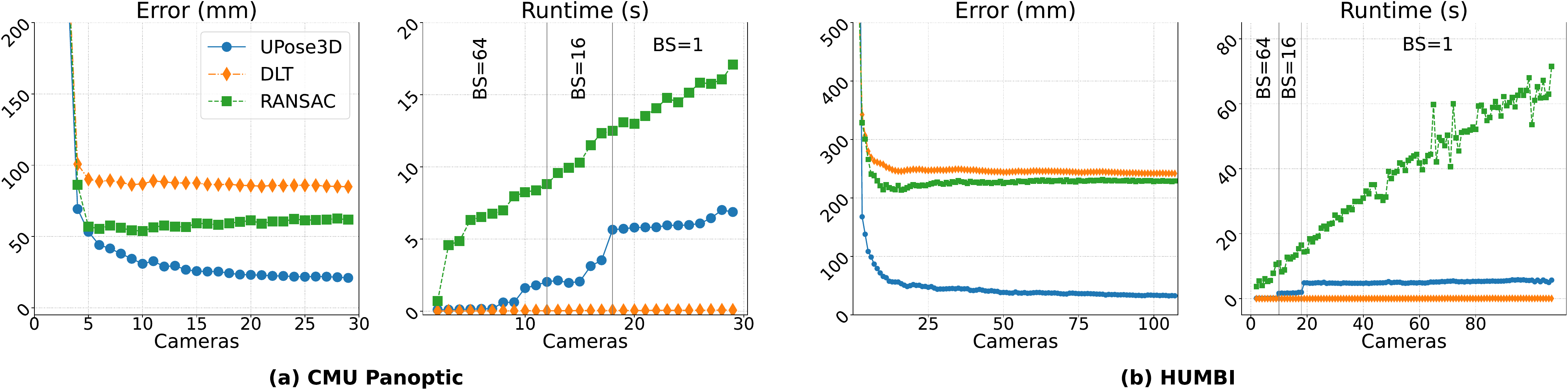}
\caption{We demonstrate the scalability of UPose3D to the number of cameras.}
\label{fig:fig4}
\end{figure}

\begin{figure}[t]
\centering
\includegraphics[width=0.85\textwidth]{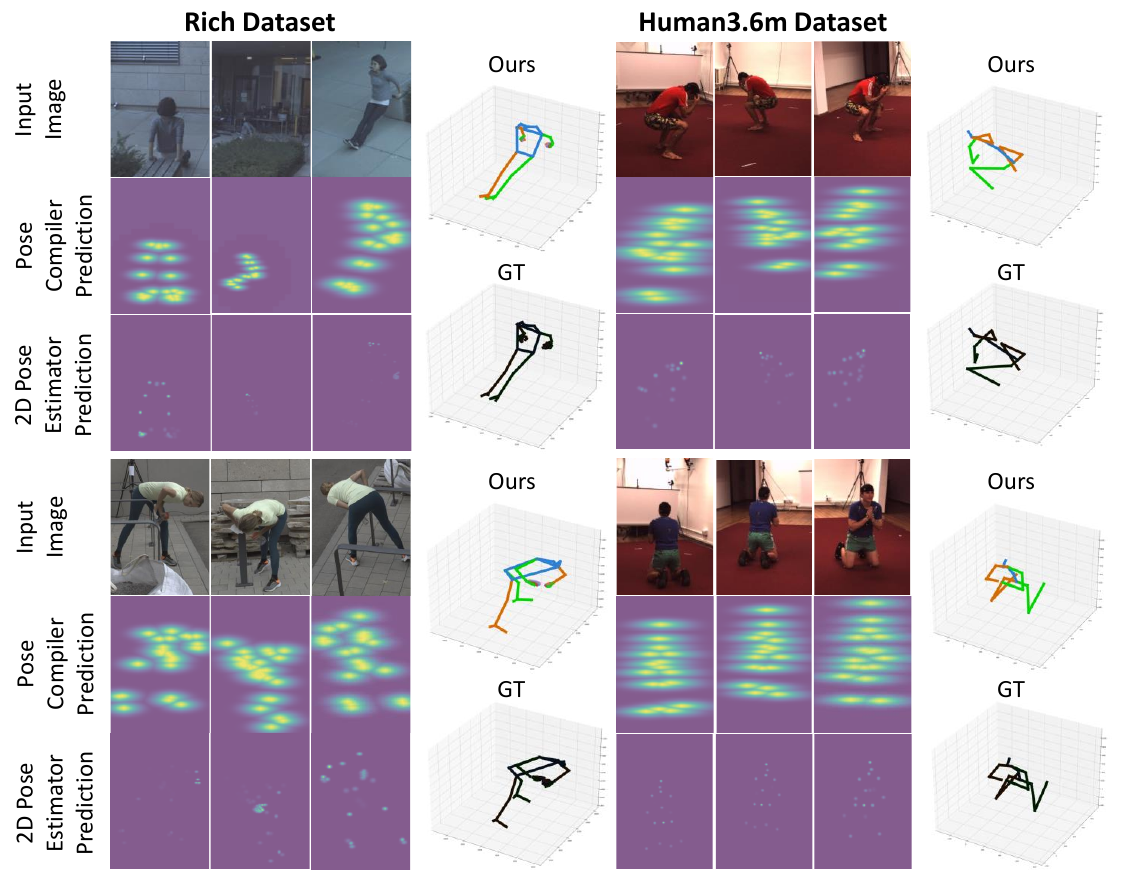}
\caption{Illustration of UPose3D on Human3.6m \cite{ionescu2013human3} (right) and RICH \cite{huang2022capturing} (left) datasets, showing the accurate 3D pose estimated by our UPose3D (top) compared to ground-truth (bottom).}
\label{fig:fig3}
\end{figure}

\subsection{Viewpoint Scalability}
In the context of multi-view 3D human pose estimation, scalability may be defined as the capability of a method to handle data from an increasing number of camera views effectively. Specifically, such a solution should have an almost constant runtime with constantly decreasing error when more cameras are used. Accordingly, to test the scalability of our method, we evaluate it with different numbers of cameras on the CMU Panoptic \cite{joo2015panoptic} and the HUMBI \cite{yoon2021humbi,yu2020humbi} datasets and present the results in \cref{fig:fig4}. For this experiment, we use the HRNet-W48 \cite{sun2019deep} backbone, similar to our experiments on the RICH dataset, to analyze scalability in OoD scenarios. Additionally, we report and compare our error and runtime with two triangulation techniques, RANSAC and DLT. We tune the RANSAC implementation of \cite{iskakov2019learnable} and run it for 20 iterations. The error analysis shows that as the number of views increases, UPose3D's error continuously improves while DLT and RANSAC reach a plateau. Next, we discard the common factor of the 2D pose estimators for runtime scalability analysis. The inference time of our pipeline with a single batch size remains constant despite the number of cameras. However, increasing the batch size causes runtime improvements at the cost of memory and runtime variability.

\subsection{Qualitative Analysis}
\cref{fig:fig3} shows the outputs of our method on challenging examples, where we sample the keypoint distributions in every pixel to demonstrate the keypoint likelihoods from the 2D pose estimator and our pose compiler as heatmaps. Despite choosing challenging scenes, we observe that our predictions are still close to the ground-truth keypoints, indicating that our method does not produce any significant outliers within its output even on the unseen samples of RICH \cite{huang2022capturing}.

\section{Conclusion}
This paper presents UPose3D, a multi-view 3D human pose estimation method designed to address the challenges in generalization, scalability, and over-reliance on real-world 3D annotated data. It includes a novel cross-view fusion strategy that scales well with varying camera numbers and volume sizes. Additionally, UPose3D integrates a pose compiler that learns to predict keypoint positions and uncertainties given the cross-view and temporal information. Finally, the uncertainties and predicted key points from two sources of the image branch and pose compiler are used for uncertainty-aware 3D pose estimation. UPose3D outperforms state-of-the-art approaches in OoD setup and achieves competitive InD results without any 3D pose annotations. Therefore, UPose3D may positively impact applications where performance and robustness are crucial.

\noindent \textbf{Future Work.}
Our primary focus in this paper was the fidelity of estimated output 3D poses from multi-view inputs without requiring 3D supervision. We acknowledge, however, that our method could be optimized for \textit{real-time} applications. A promising avenue to achieve faster inference involves the exploration of specialized second-order optimizers that use a deep-learning neural network to estimate the Hessian matrix during likelihood maximization. Furthermore, depth and trajectory modalities can be explored for additional supervision in the likelihood function to reduce noisy predictions.

\section*{Acknowledgements}
This work was performed during Vandad Davoodnia's internship at Ubisoft Laforge, partially funded by Mitacs through the Accelerate program.

\bibliographystyle{splncs04}
\bibliography{mainbib}

\clearpage

\newpage

\appendix

\section{Additional Details}
\subsection{Details on Training and Multi-view Data Synthesis}

Let $r \in \mathbb{R}^{T \times 3}$, $\Theta \in \mathbb{R}^{T \times 55 \times 3 \times 3}$, and $\beta\in\mathbb{R}^{16}$ be root position, body joint rotation matrices in a temporal window of length $T$, and shape parameters from human motion capture data. We begin our multi-view data generation by augmenting the shape parameters with Gaussian noise with a standard deviation equal to the standard deviation of all shape parameters within the datasets. Next, we apply mediolateral mirroring with a 50\% chance and randomly rotate the motion sequence around its center. We pass the augmented $\{r, \Theta, \beta \}$ parameters to the forward-kinematic layer of the SMPL body model to obtain 3D vertices. Lastly, we use a dataset-specific joint regressor on the vertices to extract the 3D keypoints used in the next steps of our pipeline.

\begin{figure*}[t]
\centering
\includegraphics[width=1.0\textwidth]{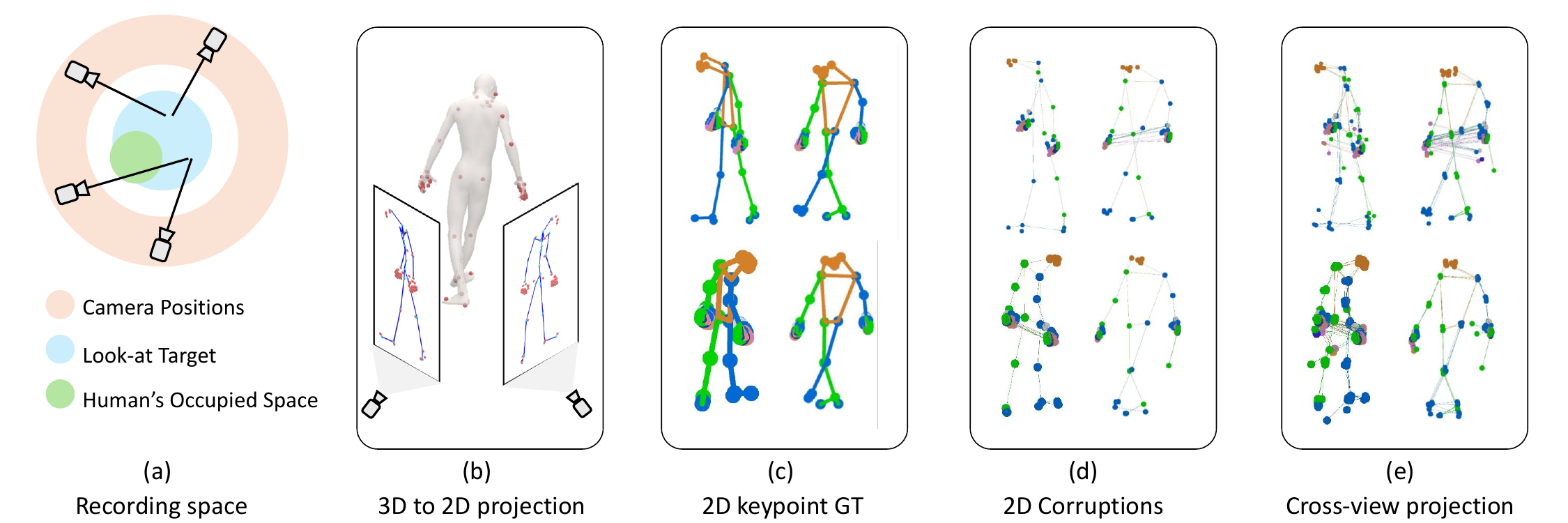}
\caption{
We illustrate our multi-view data synthesis framework, starting with (a) camera placement in a space surrounding a motion-captured human body; (b) extraction and projection of keypoints onto the synthetic cameras; (c) 2D ground-truth keypoints; (d) data corruption; and (e) cross-view projection to prepare the point cloud training data for our pose compiler.
}
\label{fig:fig5}
\end{figure*}

Next, we simulate a multi-camera recording setup by randomly positioning several cameras in cylindrical space. As depicted in \cref{fig:fig5}-a, we randomly choose a recording volume size that encircles the space occupied by the human body. To ensure that the subject appears in most cameras, we select the tilt, pitch, and yaw so that they look at a random point in the center of the recording volume while maintaining the correct up direction. Additionally, we limit the camera height to mimic typical multi-view video recording setups.

\begin{figure*}[t]
\centering
\includegraphics[width=1.0\textwidth]{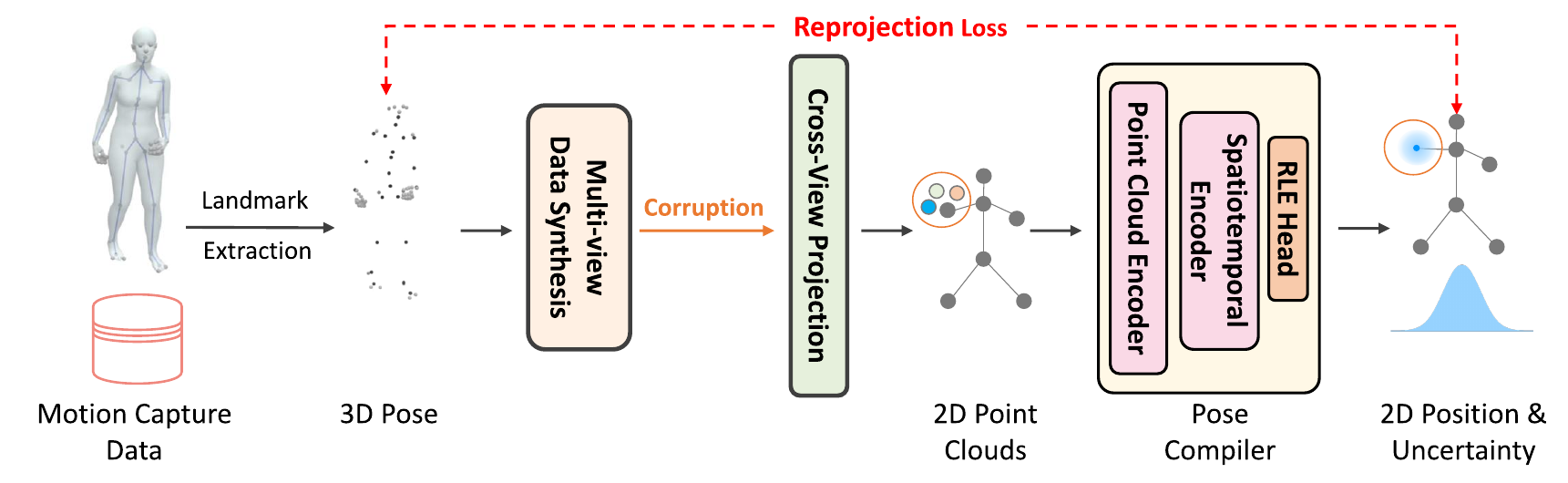}
\caption{
We illustrate the training routine of our pose compiler using synthetic data generated based on real motion capture sequences. 
}
\label{fig:fig6}
\end{figure*}

After obtaining the camera intrinsic and extrinsic parameters, we project the 3D body keypoints onto each camera view (see \cref{fig:fig5}-b). We use these 2D keypoints as ground truths $\mu_g$ to train our pose compiler (see \cref{fig:fig5}-c). We then add 2D point corruptions to the 2D keypoints, including Gaussian noise with varying standard deviations, simulated occlusions with varying sizes and probabilities, mediolateral flipping, and occasional truncation effects (see \cref{fig:fig5}-d). Next, we obtain the cross-view projected keypoints (see \cref{fig:fig5}-e) via the algorithm described in \cref{sec:meth_2_mv} in the main paper. Finally, we train our pose compiler using the ground-truth keypoints and point clouds containing noisy 2D data, as depicted in \cref{fig:fig6}.

\subsection{Details on Criss-cross Attention}
As discussed in \cref{sec:meth_3_mvf} in the main paper, we use criss-cross attention blocks in our spatiotemporal encoder to process information more efficiently. Accordingly, the cross-view input features $\{f_{i}\}$ are first projected into queries, keys, and values ($Q, K, V \in \mathbb{R}^{T \times J \times 2H}$) via a linear layer. Next, we divide them into temporal $Q_T,K_T,V_T \in \mathbb{R}^{T \times J \times H}$ and spatial groups $Q_S,K_S,V_S \in \mathbb{R}^{T \times J \times H}$. The temporal and spatial (skeleton joints) attentions are then calculated in two separate self-attention modules and concatenated before the next feed-forward layer and normalization. As a result of this operation, the receptive field of each transformer layer is the information residing on the spatial and temporal axis, and stacking multiple layers can approximate the full spatiotemporal attention without large computational overhead. In the following sections, we study the effectiveness of our design choice and compare its computation cost and performance against full attention and concurrent attention designs.

\section{Additional Experiments and Results}
This section describes the 2D datasets used during training and fine-tuning of our 2D pose estimator. We then study the details of our pipeline to evaluate its performance under different inputs, network architectures, and initialization strategies for 3D keypoint estimation. Next, we provide additional comparisons on the Human3.6m \cite{ionescu2013human3} dataset with weak or semi-supervised methods. We will also provide more comparisons with monocular pose estimation approaches on the RICH \cite{huang2022capturing} dataset.

\begin{table*}[t]
\centering
  \caption{Additional ablation study on Human3.6m dataset. We only report the computation cost of our pose compiler (in FLOPs) and exclude the CPN \cite{chen2018cascaded} network with 5.16\textit{T} FLOPs for 27 frames of 4 views. Additionally, 64.87\textit{M} of parameters in all experiments belong to the CPN network. }
  \label{tab:tab5_extra_ablation}
  \resizebox{1.0\textwidth}{!}{
  \setlength
  \tabcolsep{3pt}
  \scriptsize
  \begin{tabular}{lccccc}
    \hline    
    \textbf{Method} & \textbf{MPJPE}$\downarrow$  &\textbf{PA-MPJPE}$\downarrow$ & \textbf{Param. (M)}$\downarrow$ & \textbf{Time (s)}$\downarrow$ & \textbf{FLOPs (G)} \\
    \hline
    \hline
    \textbf{UPose3D} {\tiny (T = 27, Tol$.=10^{-3}$ \textit{mm})}    & 26.42 & 23.42  & 65.407  & 10.1  & 2.04   \\
    \ \ \ \ w/ zero init                                             & 28.51 & 32.85  & 65.407  & 12.5  & 2.04   \\
    \ \ \ \ w/ zero init {\tiny (Tol$.=10^{-6}$ \textit{mm})}        & 26.42 & 23.42  & 65.407  & 28.9  & 2.04   \\
    \ \ \ \ w/ T = 243                                                 & 33.17 & 25.11  & 62.660  & 10.9  & 20.18  \\
    \ \ \ \ w/ concurent attention                                   & 26.57 & 23.61  & 65.391  & 10.3  & 2.01   \\
    \ \ \ \ w/ full attention                                        & 26.50 & 23.57  & 65.322  & 10.3  & 2.28   \\
    \ \ \ \ w/ full attention (T = 243)                                & 34.97 & 28.60  & 65.336  & 10.3  & 51.56  \\
    \ \ \ \ w/ epipolar line                                         & 26.46 & 23.45  & 65.407  & 10.1  & 2.04   \\
    \ \ \ \ w/ relative camera pos. emb.                             & 26.37 & 23.43  & 65.407  & 10.2  & 2.04   \\
    \hline
  \end{tabular}
  }
\end{table*}

\subsection{2D Datasets}
\noindent\textbf{COCO WholeBody.} 
The COCO WholeBody \cite{jin2020whole} dataset is a large-scale whole-body pose estimation dataset with over 250K samples. This dataset is an extension of Common Objects in COntext (COCO) \cite{lin2014microsoft} dataset with the same training and testing splits. The dataset provides 133 2D keypoints (17 for body, 6 for feet, 68 for face, and 42 for hands) on in-the-wild images. We use this dataset to train our 2D pose estimator during OoD experiments.

\noindent\textbf{MPII.} 
The MPII Human Pose dataset \cite{andriluka20142d} dataset is a popular 2D pose estimation benchmark. It contains over 40,000 images of people performing over 400 actions in diverse scenarios. The dataset contains 16 body joint labels and is frequently used to pre-train \cite{zhang2021adafuse} or improve cross-dataset generalization\cite{zhou2017towards,qiu2019cross}. We use this dataset for 2D pose estimator pre-training and fine-tuning.

\subsection{Additional Ablation Study}
Following the ablation study originally presented in \cref{sec:sec_ablation} of the main paper, we investigate the impact of temporal length, our spatiotemporal encoder's architecture, different formulations of the point clouds, and our initialization strategy for 3D keypoint estimation in \cref{tab:tab5_extra_ablation}. Additionally, we report and provide the computational cost comparisons for a single input batch with T = 27 and 4 views. Our pose compiler is significantly smaller than a single 2D pose estimator, taking less than 1\% of the total parameter count.

\noindent\textbf{Random Initialization.}
We use the L-BFGS \cite{liu1989limited} optimization algorithm to solve the 3D keypoint MLE iteratively. To speed up this process, we stop the optimization when the changes of our optimization variables, namely $U$, are less than a specific tolerance (Tol$.=0.001$ \textit{mm}). We further speed up the optimization process by using a DLT algorithm to initialize the 3D points $U$. \Cref{tab:tab5_extra_ablation} first examines the effect of our initialization strategy when $U$ is initialized to zero, and the tolerance remains unchanged, showing a significant rise in the 3D keypoint estimation error and inference time. Next, \cref{tab:tab5_extra_ablation} shows that by lowering the tolerance, zero-initialization performs similarly to our proposed strategy, but at 3 times more inference time. Therefore, we conclude that unlike prior works \cite{usman2022metapose}, our method is not reliant on initialization, and the initialization only speeds up the estimation process. This may be due to the smooth nature of the uncertainty distributions learned by the normalizing flows \cite{li2021human}.

\noindent\textbf{Longer Temporal Window.}
We study the computational cost and performance impact of very long temporal context size. Following \cite{zhu2023motionbert}, we report the performance of UPose3D when 243 frames, as opposed to 27 frames, to infer the 3D keypoints of the center frame in \cref{tab:tab5_extra_ablation}. This new model takes 10 times more FLOPs to compute and does not perform as well as our original model. 
This may be because our synthetic data augmentations and corruption strategies are tuned for smaller time windows, as longer context sizes were not in our considerations. Our observations of the training and validation losses also show signs of overfitting during training for longer time windows. As extremely long context sizes are not in the scope of this paper, we do not perform any additional tuning of these models and leave them for future research.

\noindent\textbf{Pose Compiler Architecture.}
We compare the effect of our criss-cross attention modules with vanilla (full) and concurrent attention. \Cref{tab:tab5_extra_ablation} shows that criss-cross attention outperforms the other two designs while requiring less computation (FLOPs) than full attention. Additionally, we observe that on the extreme case of very long temporal context sizes (T = 243), criss-cross attention still outperforms full attention models by 1.8 \textit{mm} while requiring 60\% less computations.

\begin{table*}[t]
  \centering
  \caption{Additional comparisons with prior works on the full test set of the Human3.6m dataset in InD settings. (-) denotes that the error was not reported in the original work.}
  \label{tab:tab6_extra_h36m}
  % \resizebox{1.0\textwidth}{!}{
  \setlength
  \tabcolsep{2pt}
  \scriptsize
  \begin{tabular}{lcccccc}
    \hline    
    \textbf{Method} & \textbf{Supervision} & \textbf{Multi-view} & \textbf{Frames} & \textbf{MPJPE}$\downarrow$ & \textbf{PA-MPJPE}$\downarrow$  & \textbf{N-MPJPE}$\downarrow$  \\
    \hline
    \hline
    Rhodin \etal \cite{rhodin2018learning} & 3D           & \xmark     & 1   & 66.8   & 51.6    & 63.3  \\  
    Rhodin \etal \cite{rhodin2018learning} & Weakly 3D    & \xmark     & 1   & -      & 65.1    & 80.1  \\  
    EpipolarPose \cite{kocabas2019self}    & Weakly 3D    & \xmark     & 1   & 55.08  & 47.91   & 54.90 \\  
    CanonPose    \cite{wandt2021canonpose} & Weakly 3D    & \xmark     & 1   & -      & 53.0    & 82.0  \\  
    Gong \etal \cite{gong2022progressive}  & Synthetic 3D & \checkmark & 1   & 53.8   & 42.4    & -     \\  
    BKinD-3D \cite{sun2023bkind}           & 3D Discovery & \checkmark & 20  & 125.0  & 105.0   & -     \\  
    UPose3D (Ours)                         & 2D           & \checkmark & 1   & 26.9   & 24.1    & 26.2  \\
    UPose3D (Ours)                         & 2D           & \checkmark & 27  & 26.4   & 23.4    & 25.6  \\
    \hline
  \end{tabular}
  % }
\end{table*}

\begin{table*}[t]
\centering
  \caption{Comparison of our method in OoD setting on RICH dataset against prior works. * denotes our replication of prior works.}
  \label{tab:tab8_extra_rich}
  % \resizebox{1.0\textwidth}{!}{
  \setlength
  \tabcolsep{4pt}
  \scriptsize
  \begin{tabular}{lccccc}
    \toprule
    \textbf{Method} & \textbf{MPJPE}$\downarrow$  &\textbf{PA-MPJPE}$\downarrow$  & \textbf{OoD} & \textbf{Multi-view} & \textbf{Output} \\
    \hline
    \hline
    SA-HMR \cite{shen2023learning}        & 93.9   & -     & \xmark      & \xmark & SMPL     \\ 
    IPMAN-R \cite{tripathi20233d}         & 79.0   & 47.6  & \xmark      & \xmark & SMPL     \\ 
    METRO \cite{lin2021end}               & 98.8   & -     & \xmark      & \xmark & SMPL     \\ 
    METRO \cite{lin2021end}               & 129.6  & -     & \checkmark  & \xmark & SMPL     \\ 
    SPIN \cite{kolotouros2019learning}    & 112.2  & 71.5  & \checkmark  & \xmark & SMPL     \\ 
    PARE \cite{kocabas2021pare}           & 107.0  & 73.1  & \checkmark  & \xmark & SMPL     \\ 
    CLIFF \cite{li2022cliff}              & 107.0  & 67.2  & \checkmark  & \xmark & SMPL     \\ 
    AdaFuse\textsuperscript{*} \cite{zhang2021adafuse}                     & 524.0 & 85.8 & \checkmark  & \checkmark & 3D Keypoints \\ 
    HRNet-W48\Plus Grid Search\textsuperscript{*}                               & 64.4  & 54.9 & \checkmark  & \checkmark & 3D Keypoints \\ 
    HRNet-W48\Plus DLT\textsuperscript{*}                                  & 66.0  & 55.1  & \checkmark  & \checkmark & 3D Keypoints \\ 
    Ours ($T=1$)                                         & 36.2  & 33.4  & \checkmark  & \checkmark & 3D Keypoints \\ 
    Ours ($T=27$)                                        & 34.7  & 32.0  & \checkmark  & \checkmark & 3D Keypoints \\ 
    \hline 
  \end{tabular}
  % }
\end{table*}

\noindent\textbf{Inputs of Pose Compiler.}
Finally, we investigate the effect of different point cloud formation strategies in our pipeline. Specifically, we study the impact of appending a relative camera position embedding, inspired by \cite{shuai2022adaptive}, to the cross-projected 2D keypoints while creating the point clouds. Accordingly, in our first experiment, we concatenate the epipolar line parameters of other views to the point cloud of the reference view. Similarly, in our second experiment, we concatenate the relative position of the other cameras to the input point cloud as well. However, as shown in \cref{tab:tab5_extra_ablation}, adding extra inputs does not greatly impact the performance.

\subsection{Additional Baselines and Comparisons}
\Cref{tab:tab6_extra_h36m} complements \cref{tab:tab1_h36m} of the main paper by providing more comparisons with prior works on the Human3.6m \cite{ionescu2013human3} dataset. Here, we include 3D keypoint estimation approaches regardless of their input modality or supervision type in InD settings. We observe that our method outperforms all of the other approaches despite only using 2D supervision. Additionally, in \cref{tab:tab8_extra_rich}, we compare our work with prior research on the RICH \cite{huang2022capturing} dataset. Since this dataset was recently published, only monocular 3D body modeling techniques have reported their performance on this dataset. Here, we observe that our method outperforms the majority of prior works. More importantly, when comparing \cref{tab:tab6_extra_h36m} and \cref{tab:tab8_extra_rich} we notice that our method achieves consistent results between InD and OoD evaluations on the Human3.6m \cite{ionescu2013human3} and the RICH \cite{huang2022capturing} datasets, showing generalizability across in-studio and outdoor environments.

\section{Additional Visual Examples}
We provide a supplementary video that describes our method with visual demonstrations. Additionally, we provide several video clips of input and output data from Human3.6m \cite{ionescu2013human3}, RICH \cite{huang2022capturing}, and CMU-Panoptic \cite{joo2015panoptic} datasets and compare the visual fidelity of our approach with the state-of-the-art method on Human3.6m, AdaFuse \cite{zhang2021adafuse}.

\section{Qualitataive Comparisons}
\begin{figure}[t]
\centering
\includegraphics[width=0.9\textwidth]{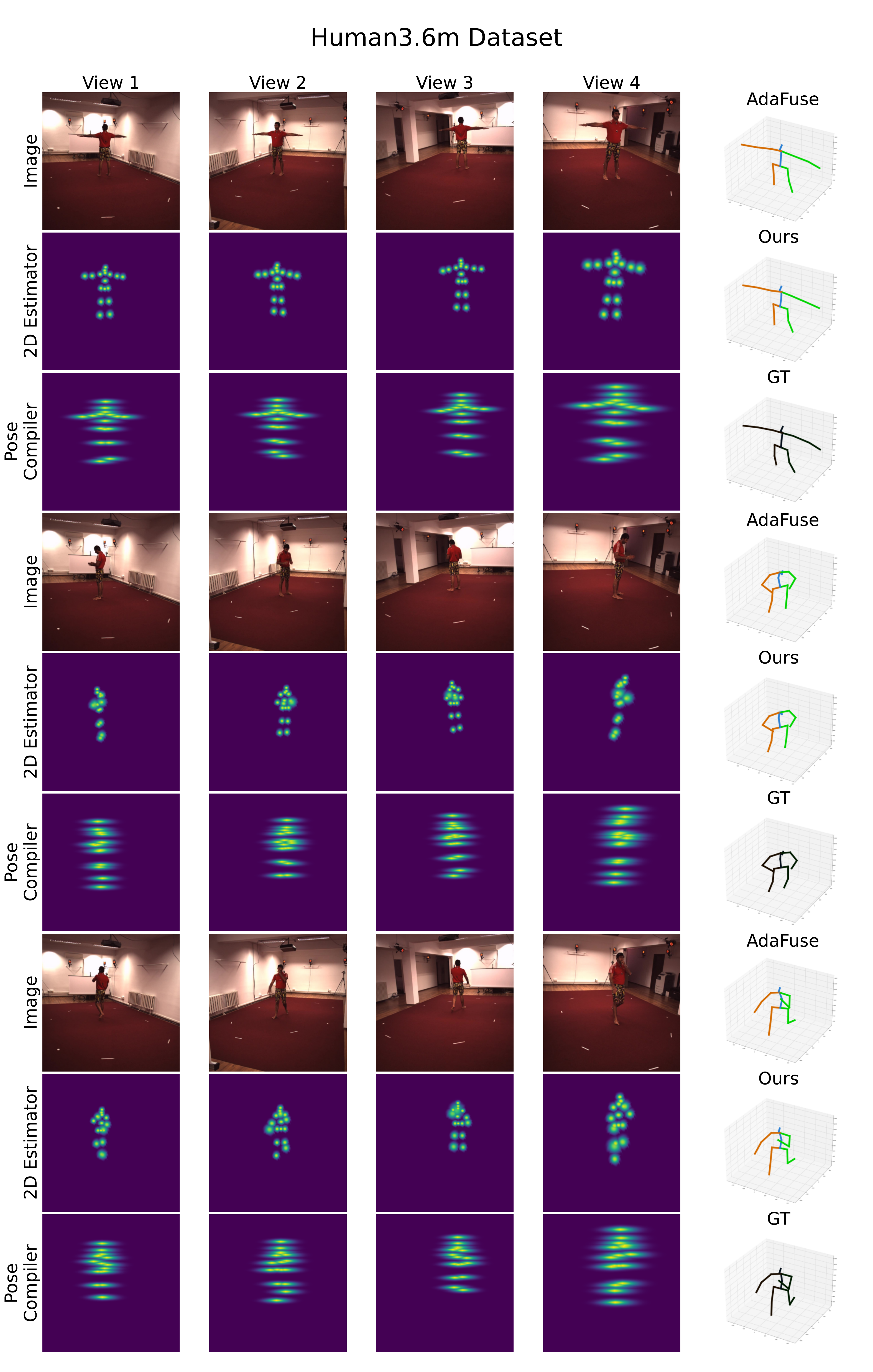}
\caption{
Example output of our proposed UPose3D pipeline in comparison to AdaFuse \cite{zhang2021adafuse} is presented in the InD evaluation scheme on the Human3.6m \cite{ionescu2013human3} dataset.
}
\label{fig:fig7}
\end{figure}

\begin{figure} 
\centering
\includegraphics[width=0.9\textwidth]{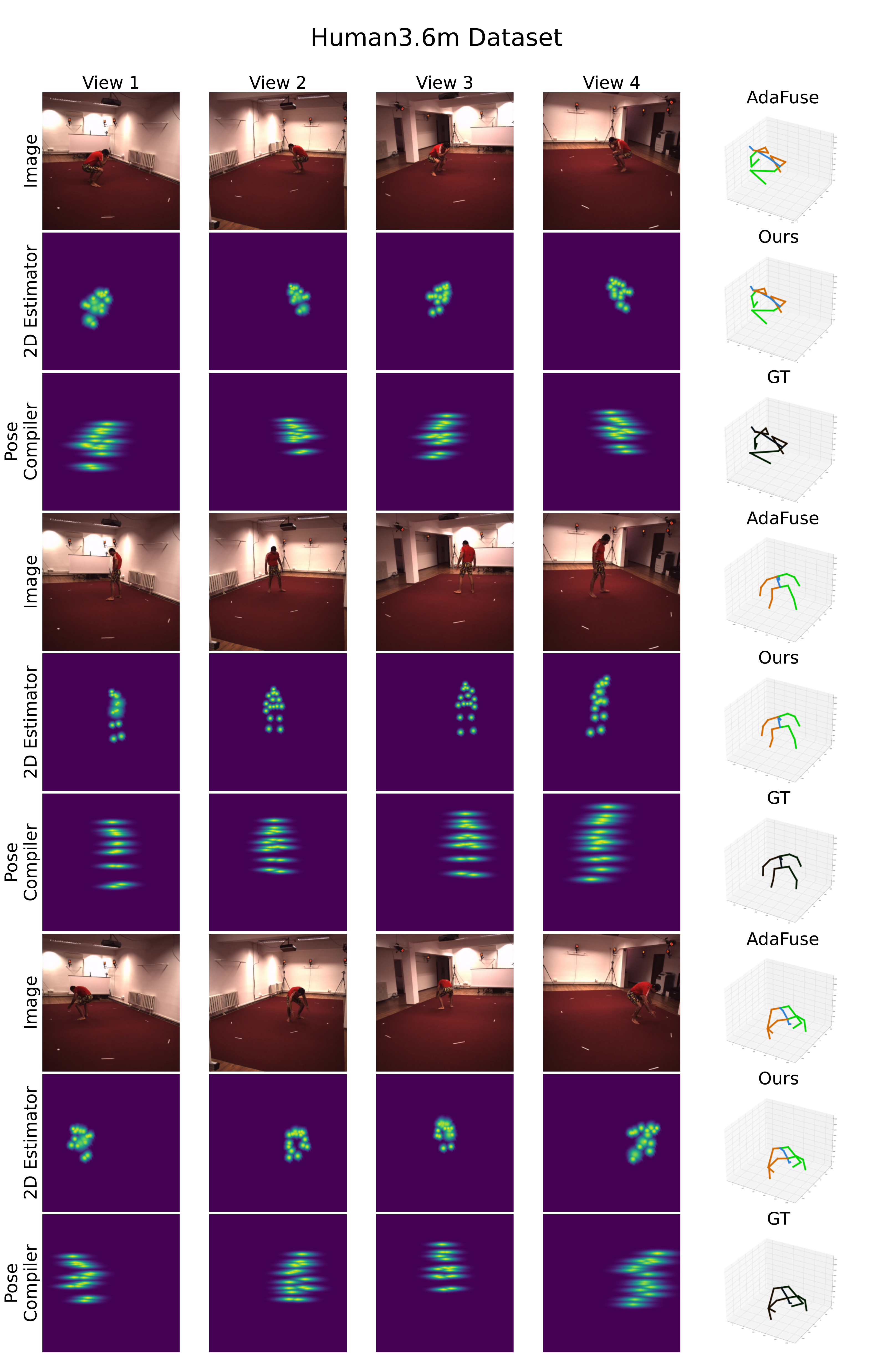}
\caption{
Example output of our proposed UPose3D pipeline in comparison to AdaFuse \cite{zhang2021adafuse} is presented in the InD evaluation scheme on the Human3.6m \cite{ionescu2013human3} dataset.
}
\label{fig:fig8}
\end{figure}

\begin{figure}
\centering
\includegraphics[width=0.9\textwidth]{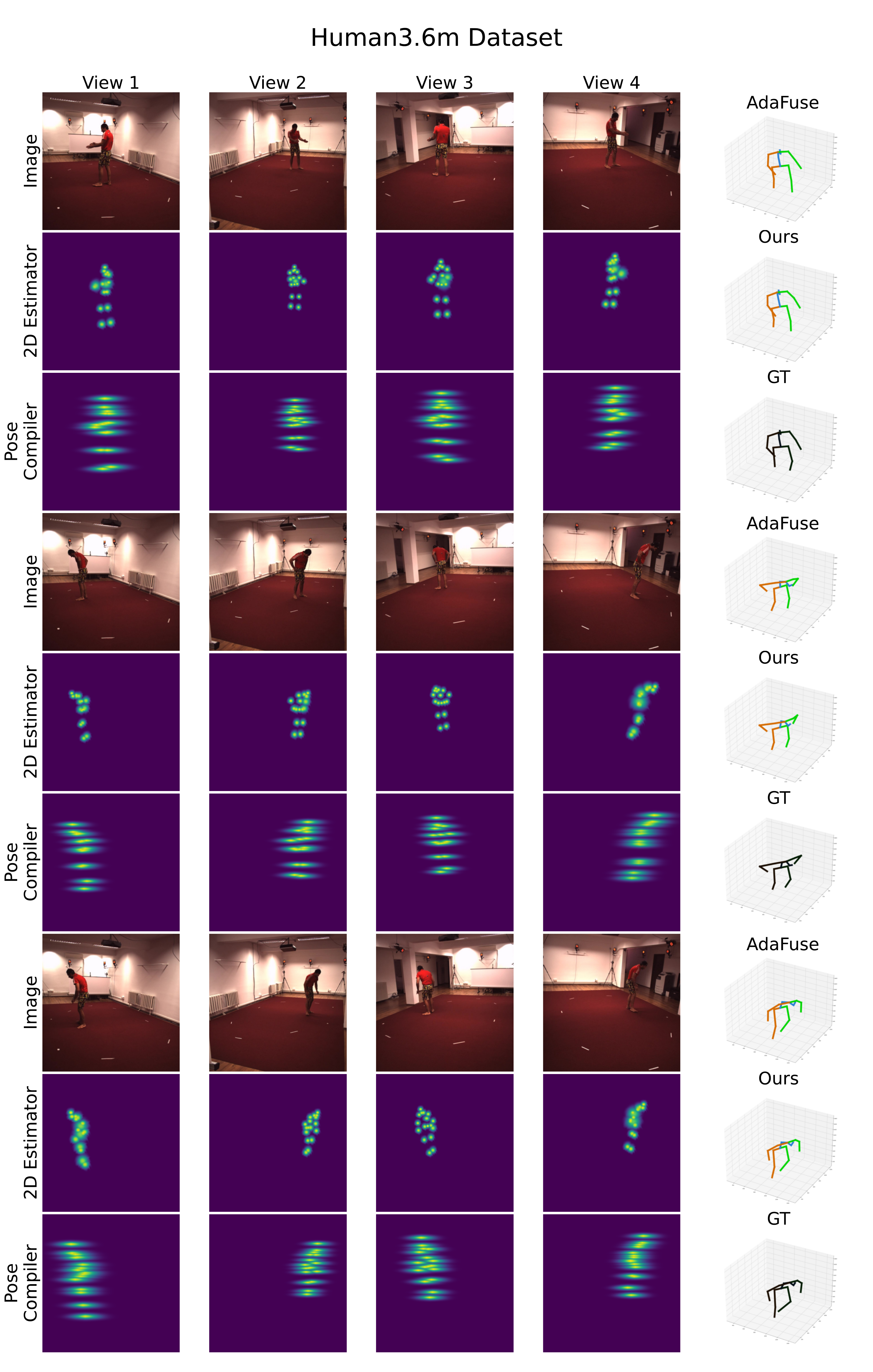}
\caption{
Example output of our proposed UPose3D pipeline in comparison to AdaFuse \cite{zhang2021adafuse} is presented in the InD evaluation scheme on the Human3.6m \cite{ionescu2013human3} dataset.
}
\label{fig:fig9}
\end{figure}

\begin{figure} 
\centering
\includegraphics[width=0.9\textwidth]{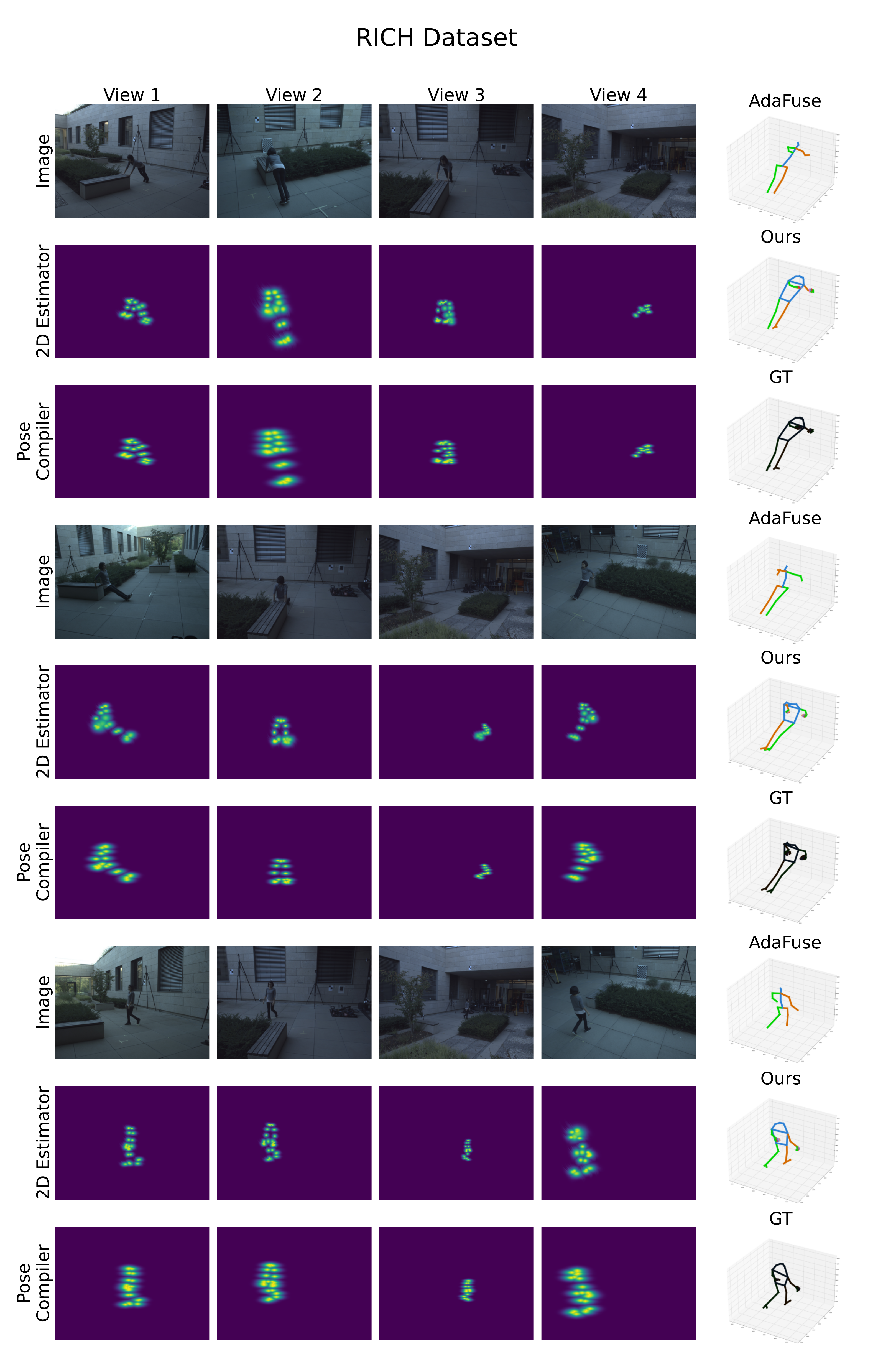}
\caption{
Example output of our proposed UPose3D pipeline in comparison to AdaFuse \cite{zhang2021adafuse} is presented in the OoD evaluation scheme on the RICH \cite{huang2022capturing} dataset. The first and second samples show the effectiveness of our approach in solving occlusions for detecting hands and feet.
}
\label{fig:fig10}
\end{figure}

\begin{figure} 
\centering
\includegraphics[width=0.9\textwidth]{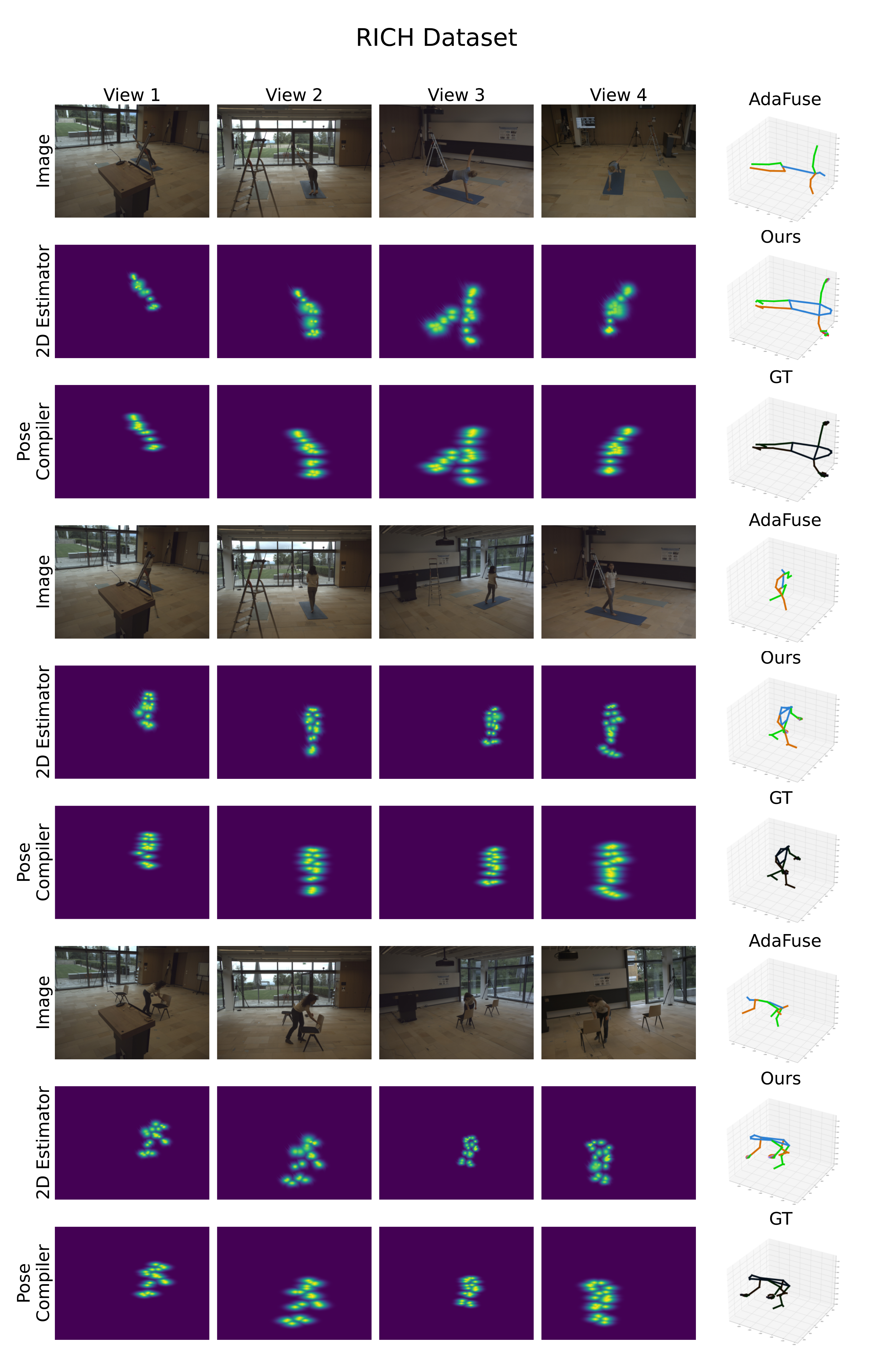}
\caption{
Example output of our proposed UPose3D pipeline in comparison to AdaFuse \cite{zhang2021adafuse} is presented in the OoD evaluation scheme on the RICH \cite{huang2022capturing} dataset. The first sample illustrates a challenging input with a rare posture, where both AdaFuse and our method successfully predict the correct posture.
}
\label{fig:fig11}
\end{figure}

\begin{figure} 
\centering
\includegraphics[width=0.9\textwidth]{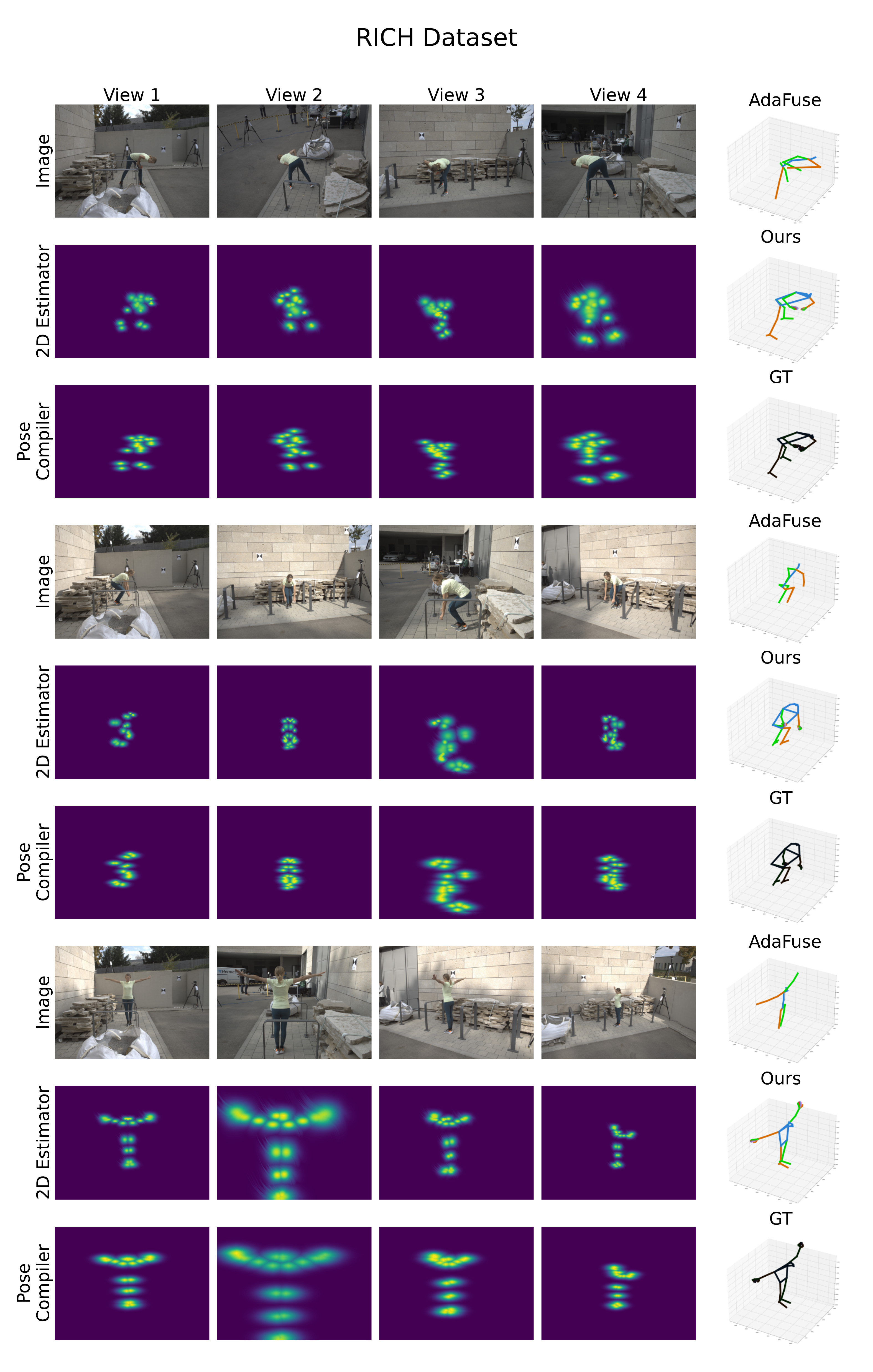}
\caption{
Example output of our proposed UPose3D pipeline in comparison to AdaFuse \cite{zhang2021adafuse} is presented in the OoD evaluation scheme on the RICH \cite{huang2022capturing} dataset. We observe that our method outperforms AdaFuse in the first and third samples.
}
\label{fig:fig12}
\end{figure}

In \cref{fig:fig7}, \cref{fig:fig8}, and \cref{fig:fig9}, we demonstrate some examples of our UPose3D on Human3.6m \cite{ionescu2013human3} dataset to showcase its visual fidelity in comparison to ground-truth keypoints and AdaFuse \cite{zhang2021adafuse} in InD evaluation scheme. Additionally, we provide more visual examples of UPose3D results in \cref{fig:fig10}, \cref{fig:fig11}, and \cref{fig:fig12} in comparison to our implementation of AdaFuse \cite{zhang2021adafuse} in OoD settings on the RICH \cite{huang2022capturing} dataset. To better visualize the sharp keypoint distribution output of our 2D pose estimators, we show the logarithm of heatmaps in all figures for 2D pose estimators. We refer the reader to Fig. 4 of the main paper for an illustration of the real heatmaps without any post-processing. Our method performs consistently in both settings, while AdaFuse fails to correctly predict the human keypoints in some OoD samples. In all cases, the 2D pose estimator generally results in more refined predictions and sharper uncertainty distributions, while our pose compiler outputs a coarser distribution. Moreover, our method typically depicts higher horizontal uncertainties, which may be due to more frequent horizontal movements.

\end{document}